\pdfoutput=1

\documentclass[11pt]{article}

\usepackage[]{emnlp2021}

\usepackage{times}
\usepackage{latexsym}

\usepackage{amsmath,amsfonts,amssymb,commath}
\usepackage{booktabs,pgfplots,adjustbox}
\usepackage{xcolor}
\newcommand{\bftab}{\fontseries{b}\selectfont}
\usepackage[normalem]{ulem}
\useunder{\uline}{\ul}{}
\usepackage{placeins}

\usepackage[T1]{fontenc}

\usepackage[utf8]{inputenc}

\usepackage{microtype}

\title{Learning Zero-Shot Multifaceted Visually Grounded Word Embeddings via Multi-Task Training}

\author{Hassan Shahmohammadi \quad Hendrik P. A. Lensch \quad R. Harald Baayen \\ \\
        University of Tübingen\\ \{hassan.shahmohammadi, hendrik.lensch, harald.baayen\}@uni-tuebingen.de}
        
\begin{document}

\maketitle

\begin{abstract}
Language grounding aims at linking the symbolic representation of language (e.g., words) into the rich perceptual knowledge of the outside world.  The general approach is to embed both textual and visual information into a common space -the grounded space- confined by an explicit relationship. We argue that since concrete and abstract words are processed differently in the brain, such approaches sacrifice the abstract knowledge obtained from textual statistics in the process of acquiring perceptual information.
The focus of this paper is to solve this issue by implicitly grounding the word embeddings. Rather than learning two mappings into a joint space, our approach integrates modalities by implicit alignment. This is achieved by learning a reversible mapping between the textual and the grounded space by means of multi-task training.
Intrinsic and extrinsic evaluations show that our way of visual grounding is highly beneficial for both abstract and concrete words. Our embeddings are correlated with human judgments and outperform previous works using pretrained word embeddings on a wide range of benchmarks. Our grounded embeddings are publicly available \href{https://github.com/Hazel1994/Visually_Grounded_Word_Embeddings}{here}.
\end{abstract}

\section{Introduction}
The distributional hypothesis asserts that words occurring in similar contexts are semantically related \citep{harris1954distributional}.  Current state-of-the-art word embedding models \citep{pennington-etal-2014-glove,peters-etal-2018-deep}, despite their successful application to various NLP tasks \citep{wang-etal-2018-glue}, suffer from the lack of grounding in general knowledge  \citep{harnad1990symbol, burgess2000theory}, such as captured by human perceptual and motor systems \citep{pulvermuller2005brain, therriault2009role}. To overcome this limitation, research has been directed to linking word embeddings to perceptual knowledge in visual scenes. Most studies have attempted to bring visual and language representations into close vicinity in a common feature space \citep{silberer-lapata-2014-learning,kurach2017better, kiela-etal-2018-learning}. 

However, studies of human cognition indicate that the brain processes abstract and concrete words differently \citep{paivio1990mental,anderson-etal-2017-visually} due to the difference in associated sensory perception. According to \citet{montefinese2019semantic}, similar activity for both categories are observed in the perirhinal cortext, a region related to memory and recognition, whereas in the parahippocampal cortex, associated with memory formation, higher activity only occurs for abstract words. 


We argue that forcing the textual and visual modalities to be represented in a shared space causes grounded embeddings to suffer from the bias towards concrete words as reported by  \citet{park-myaeng-2017-computational,kiela-etal-2018-learning}. Therefore, we propose a zero-shot approach that implicitly integrates perceptual knowledge into pre-trained textual embeddings (GloVe \citep{pennington-etal-2014-glove} and fastText \citep{bojanowski-etal-2017-enriching}) via multi-task training. Our approach learns multifaceted grounded embeddings which capture multiple aspects of words' meaning and are highly beneficial for both concrete and abstract words.

\begin{figure*}[ht]
  \includegraphics[width=1\textwidth]{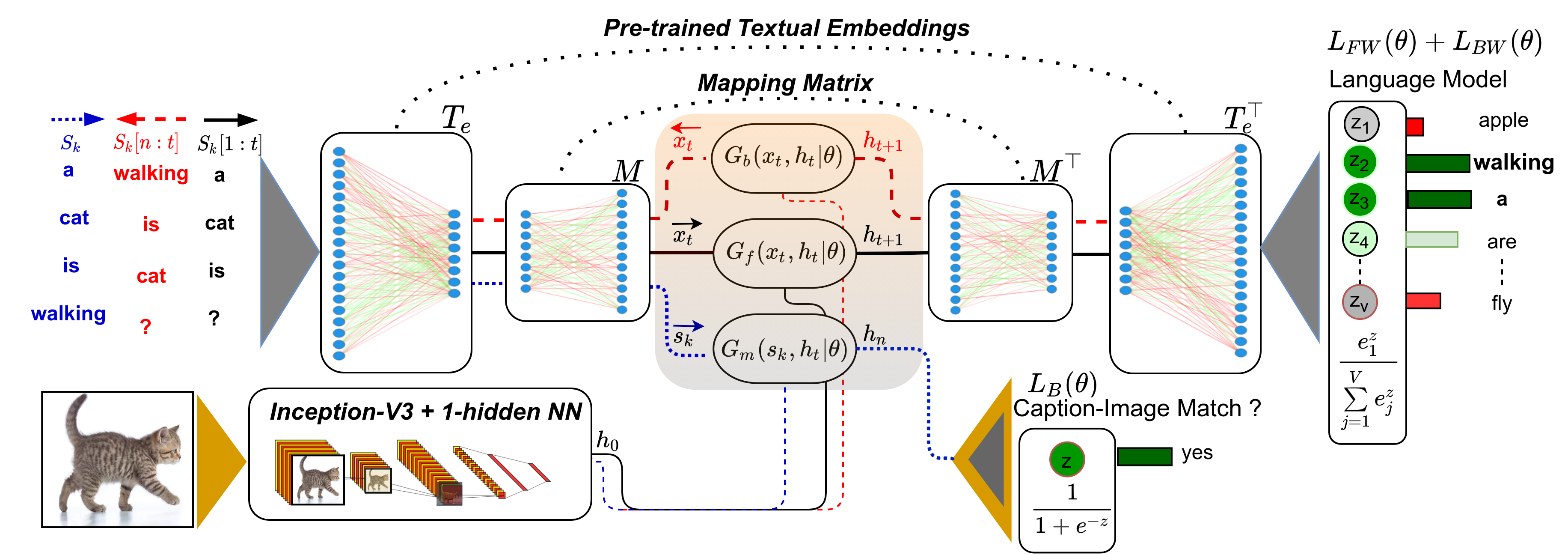}
  \caption{ \label{fig:model} Our zero-shot model: 1. Two GRU based language-model tasks in forward ($G_f$) and backward ($G_b$) directions represented by solid black and dashed red lines. 2. A matching task predicting if the given (sentence, image) pair match (blue dotted line). The zero-shot mapping matrix $M$, shared by all the tasks, learns to visually ground the textual word vectors by learning a reversible mapping from textual space to grounded space.} 
  \end{figure*}

Figure~\ref{fig:model} lays out the architecture of our model. It learns a reversible mapping from pre-trained text-based embeddings to grounded embeddings which maintains the linguistic co-occurrence statistics while integrating visual information. The architecture features a similar structure as an auto-encoder \citep{press-wolf-2017-using} translating from words to grounded space and back. 
The training is carried out as multi-task learning by combining image captioning in two directions and image-sentence pair discrimination.
At the core is a mapping matrix that acts as an intermediate representation between the grounded and textual space, which learns to visually ground the textual word vectors. 
This mapping is trained on a subset of words and then is applied to ground the full vocabulary of textual embeddings in a zero-shot manner.

We evaluate our grounded embeddings on both intrinsic and extrinsic tasks \citep{wang2019evaluating} and show that they outperform textual embeddings and previous related works in the majority of cases.
Overall, our contributions are the following:\\ a) we design a language grounding framework that can effectively ground different pre-trained word embeddings in a zero-shot manner;\\ b) we create visually grounded versions of two popular word embeddings and make them publicly available;\\ c) unlike many previous works, our embeddings support both concrete and abstract words;\\ d) we show that visual grounding has the potential to refine the irregularities of a  text-based vector space.

\section{Related Works}
The many attempts to combine images and text in order to obtain visually grounded word/sentence representations can be grouped into the following categories. 

\noindent \textbf{Feature Level Fusion:} where the grounded embedding is the result of combining the visual and textual features. Combining strategies range from simple concatenation to adopting SVD and GRU gating mechanisms  \citep{bruni2014multimodal,kiela-bottou-2014-learning, kiros-etal-2018-illustrative}. 

\noindent \textbf{Mapping to Perceptual Space:} this is usually a regression task predicting the image vector given its corresponding textual vector. The grounded embedding are extracted from an intermediate layer in auto-encoders  \citep{silberer-lapata-2014-learning,hasegawa-etal-2017-incorporating}, the output of an MLP \cite{collell2017imagined} or an RNN \citep{kiela-etal-2018-learning}. Another method is mapping both modalities into a common space in which their distance is minimized \citep{kurach2017better, park-myaeng-2017-computational}.  

\noindent \textbf{Equipping Distributional Semantic Models with Visual Context:} here images are treated as a context in the process of computing the word vectors. Many of these approaches modify the Word2Vec \citep{mikolov2013distributed} and GloVe \citep{pennington-etal-2014-glove} models by incorporating image features to the context for concrete words \citep{hill-korhonen-2014-learning,kottur2016visual,zablocki2017learning,ailem-etal-2018-probabilistic}; minimizing the max-margin loss between the image-vector and its corresponding word vectors \citep{lazaridou-etal-2015-combining}; providing social cues based on child-directed speech along with visual scenes \citep{lazaridou-etal-2016-multimodal}; or by extracting the relationship between words and images using multi-view spectral graphs \citep{fukui-etal-2017-spectral}.

\noindent \textbf{Hybrid:} this category covers the combination of previous methods and other strategies. Here, the grounded word vectors are usually the results of updating the textual word vectors during training \citep{mao2016training} or the output of sentence encoders such as LSTM \citep{hochreiter1997long}. Such methods include predicting the image vector along with training a language model \citep{chrupala-etal-2015-learning} or generating an alternative caption at the same time \citep{kiela-etal-2018-learning}. Other approaches such as using the coefficients of classifiers for grounded representation have also emerged \citep{moro2019composing}. Our model falls in the hybrid category as we take a multitasking approach. However, unlike some previous works \citep{kiela-etal-2018-learning, collell2017imagined, bordes-etal-2019-incorporating} we do not impose explicit constraints between the image features and their captions.
Our model learns the relationship indirectly via multi-task training.

\section{Multi-Task Visual Grounding}
\label{sec:approach}

In this section, we present the details of the developed method. The training data set $D$ consists of image--caption pairs, \((S_k,I_k) \in D\), with  \(S_k=[w_1,w_2...w_n]\) being a sentence with $n$ words describing the image $I_k$. 
We use the Microsoft\_COCO\_2017 dataset \citep{lin2014microsoft} in our experiments.
Let $T_e(w)\in \mathbb{R}^d$ be a pre-trained textual embedding of the word $w$, which has been trained on textual data only (e.g., GloVe). The objective is to train a mapping matrix ${M}$ to ground the word vector $T_e(w)$ visually, resulting in a grounded embedding $G_e(w) =T_e(w) \cdot M $, where $ G_e(w)\in \mathbb{R}^c$. 
To do so, we train the matrix $M$ to refine the textual vector space via two image-based language model tasks and a binary discrimination task on image-sentence pairs. For the language models, a GRU \citep{cho2014learning} is trained to predict the next word, given the previous words in the sentence provided as image caption, and its associated image vector. The transpose of the textual embedding $T_e$ is used to compute the probability distribution over the vocabulary (see Figure\ref{fig:model}).
We employ an identical scenario to form a second language model task using another GRU, where the sentence is fed backward into the model.

The image-sentence discrimination is a binary classification task predicting whether the given sentence $S_k$ represented in the grounded space matches the image $I_k$.
By training the model simultaneously on these three tasks confined by a linear transformation, we augment the visual information into the grounded embeddings (output of mapping matrix in Figure~\ref{fig:model}) while preserving the underlying structure of the textual embeddings.

\subsection{Language Model}
\label{sec:languageModel}
Given the input caption associated with image $I_k$ as  $S_k=[w_1,w_2...w_n]$, we first encode the words using a pre-trained textual embedding $T_e$ to obtain the embeddings as $S_t=[t_1,t_2...t_n]$. We then linearly project these embeddings from the textual space into the visually grounded space via the trainable mapping matrix $M$ as
$G_e(S_k) = S_t \cdot M$, 
to obtain a series of grounded vectors $G_e(S_k) =[x_1,x_2...x_n]$ where $x_i\in \mathbb{R}^c$. 
In the grounded space, the perceptual information of the image $I_k$ corresponding to $S_k$ is fused using a single-layer GRU ($G_f$ ($f$--forward) in Figure~\ref{fig:model}) that predicts the next output 
$ h_{t+1} = GRU_f(x_t,h_t|\theta)$, where $\theta$ denotes the trainable parameters, $x_t$ the current input ($G_e(w_t)$), and $h_t \in \mathbb{R}^c$ the current hidden state.

Image information is included by initializing the first hidden state $h_0$ with the image vector of $I_k$. The GRU update gate propagates perceptual knowledge from images into the mapping matrix. This has been shown to be more effective than providing the image vector at each time step as input \citep{mao2016training}.

The transpose of the mapping matrix ($M^{\top}$) is used to map back from grounded space to the textual space. That is, the output of the GRU in each time-step is mapped back into the textual space as $w_{next}= h_t \cdot M^{\top}$, 
where $w_{next} \in \mathbb{R}^d$ is an approximation of the next word's textual embedding. The mapping matrix $M$ is used to both encode and decode into/from the grounded space. This improves generalization \citep{press-wolf-2017-using} and prevents the vanishing gradient problem compared to the case where the mapping matrix is only used at the beginning of the network \citep{mao2016training}.
$w_{next}$ is fed into the transpose of the textual embeddings in the same scenario: $z = T^{\top}_e(w_{next})$, where
$z \in \mathbb{R}^{\abs{V}}$ and $V$ indicates the vocabulary. The final probability distribution over $V$ is computed by a softmax:
\begin{equation}
P(y=j | z) = \frac{e^{{z}^\top W_j}}{\sum_{c=1}^{V} e^{z^{\top}W_c}  }
\end{equation}
Defining the input (previous words and the image vector) and the predicted output (next word prediction) as above, we minimize the categorical cross entropy which is computed for batch $B$ as:
\begin{equation}
 \mathcal{L}_{FW}(\theta) =  -\frac{1}{\abs{B}}\sum_{i \in B}  \sum_{c \in V} y_{i,c} log(\hat{y}_{i,c}),
\end{equation}
Where $\hat{y}_{i,c}$ and $y_{i,c}$ are the predicted probability and ground truth for sample $i$ with respect to the class $c$.

Moreover, we define a second similar task:
Given the input caption associated with image $I_k$ as  $S_k=[w_1,w_2...w_n]$, we reverse the order of the words: 
$S_k=[w_n,w_{n-1}...w_1]$ and use another GRU ($G_b$ ($b$--backward) in Figure~\ref{fig:model}) with identical structure trained on the loss  $ \mathcal{L}_{BW}(\theta)$. The rest of the network is shared between these two tasks. 
Having this backward language model is analogous to bi-directional GRUs  \citep{schuster1997bidirectional} which, however, can not be used directly since the ground truth would be exposed by operating in both directions.

\subsection{Image-sentence discrimination}
\label{sec:binary}

Even though context-driven word representations are a powerful way to obtain word embeddings \citep{pennington-etal-2014-glove,peters-etal-2018-deep}, the performance of such models varies on language-vision tasks \citep{burns2019language}. Therefore, we propose yet another task to align the textual word vectors to their real-world relations in the images. The discrimination task predicts whether the given image and sentence describe the same content or not (shown by `caption-image match?' in Figure~\ref{fig:model}). These types of tasks have been shown effective for learning cross-modality representations \citep{lu2019vilbert, tan2019lxmert}.

Given the input caption for image $I_k$ as $S_k=[w_1,w_2...w_n]$, after projecting the embeddings into the grounded space as before, we encode the whole sentence by employing a third single-layer GRU ($G_m$ in Figure~\ref{fig:model}) with the same structure as before $h_n = GRU_m(G_e(S_k),h_0|\theta)$. Where the last output $h_n$ encodes the whole sentence. $h_0$ is again initialized with the image vector of $I_k$.
The final output is computed by a sigmoid function. This task shares the mapping matrix $M$ and the textual embeddings $T_e$. We minimize the binary cross entropy, which could be computed for each batch as:
\begin{equation} 
\mathcal{L}_{B}(\theta) =  -\frac{1}{\abs{B}}\sum_{i \in B} y_{i} log(\hat{y}_{i})  + (1-y_{i})log(1-\hat{y}_{i})
\end{equation}
For negative mining, half of the captions in each batch are replaced with captions of different, random images.

\subsection{Regularization and overall loss}
\label{sec:loss}
All the three tasks explained above share the pre-trained textual embeddings (see Figure~\ref{fig:model}) which gives rise to the question of whether the textual embeddings should be updated or kept fixed during training. By updating, we might distort the pre-trained semantic relations, especially given our limited training data. Keeping them fixed, on the other hand, does not provide the flexibility to generate the desired grounding as these embeddings are noisy and not perfect \citep{yu-etal-2017-refining}. To prevent distorting the semantic information of words while retaining sufficient flexibility, we propose the following regularization on the embedding matrix $T_e$:
\begin{equation}
\mathcal{R}(\alpha,\beta) = \frac{\alpha}{\abs{V}} \sum_{w \in V}{ \abs{ \beta - \frac{w_e \cdot w_n}{ \norm{w_e} \norm{w_n}} }},
\end{equation}
where $\alpha$ controls the overall impact and $\beta$ controls how much the new word vectors $w_n$ are allowed to deviate from the pre-trained embedding $w_e$. $\beta = 1$ indicates no deviation and $\beta = 0$ allows for up to $90$ degree deviation from $w_n$ when minimizing the equation.
We join all the tasks into a single model and minimize the following loss:
\begin{equation}
\mathcal{L}_{All}(\Theta)  = \mathcal{L}_{FW}(\theta) + \mathcal{L}_{BW}(\theta) + \mathcal{L}_{B}(\theta) + \mathcal{R}(\alpha,\beta) 
\end{equation}
where $\Theta$ denotes all the trainable parameters.

\section{Experimental setup}
\label{sec:experiments}
We use the Microsoft\_COCO\_2017 dataset \citep{lin2014microsoft} for training. Each sample contains an image with 5 captions. The dataset is split into $118k$ train and $5k$ validation samples. Each batch includes $256$ image vectors along with one of their captions. Hence, multiple image vectors might occur in each batch. Image vectors are obtained by transferring the penultimate layer of pre-trained Inception-V3 \citep{szegedy2016rethinking} trained on ImageNet \citep{deng2009imagenet}. A NN with one hidden layer and $tanh$ activation is employed to project the image vectors into the initial hidden state of the GRUs: $h_t \in \mathbb{R}^{1024}$. 
We lowercase all the words, delete the punctuation marks, and only keep the top $10k$ most frequent words. Two popular pre-trained textual word embeddings namely GloVe ($crawl-300d-2.2M-cased$) and fastText ($crawl-300d-2M-SubW$) are used for initialization of the embedding $T_e$. The mapping matrix $M$  transforms the textual embeddings into the grounded space. We investigate the best dimension of this step and the improvement over pure textual embeddings in the next sections. 
Batch normalization \citep{ioffe2015batch} is applied after each GRU. 
For the regularization, $\mathcal{R}(\alpha =0.001,\beta = 1)$ for GloVe and $\mathcal{R}(\alpha =0.01,\beta = 0)$ for fastText yielded the best relative results by meta parameter search. This shows that FastText embeddings require more deviation ($\beta = 0$ indicates $90$ degree deviation) to adapt to the proposed tasks. We trained the model for $20$ epochs with $5$ epochs tolerance early stopping using NAdam \citep{dozat2016incorporating} with a learning rate of $0.001$.

As we train a single mapping matrix $M$ for projecting from textual to grounded space, it can be used after the training to transfer out-of-vocabulary (OoV) word-vectors into the grounded space in a zero-shot manner. This way, visually grounded versions of both Glove and fastText are obtained despite being exposed to only $10k$ words.

\section{Evaluations}

\label{sec:evaluations}
While the question of what is a good word embedding model is still open \citep{wang2019evaluating}, there are two main categories of evaluation methods: intrinsic and extrinsic. Intrinsic evaluators measure the quality of word embeddings independent of any downstream tasks. For instance, quality can be assessed by comparing similarities between embeddings with word similarities as perceived by human raters.
%
Extrinsic evaluators on the other hand assess the performance based on sentence-level downstream tasks. There is not necessarily a positive correlation between intrinsic and extrinsic methods for a word embedding model \citep{wang2019evaluating}. Nonetheless, we use both types of evaluators to compare our visually grounded embeddings with those presented in related works as well as to purely text-based embeddings.\\
\textbf{Baselines:} we considered two types of embeddings as baselines 1) the pre-trained textual embeddings $T_e$, 2) $T_e$ refined based only on the captions without injecting any image information using a similar language modeling task $\mathcal{L}_{FW}$ with a one-layer GRU ($h_t \in \mathbb{R}^{1024}$ ) followed by a fully connected layer.
We refer to the second baseline as C\_GloVe and C\_fastText for Glove and fastText trained only on captions.\\
\textbf {Intrinsic Evaluators:} We evaluate on some of the common lexical semantic similarity benchmarks:  MEN \citep{bruni2014multimodal}, SimLex999 \citep{hill2015simlex}, Rare-Words \citep{luong-etal-2013-better}, MTurk771 \citep{halawi2012large}, WordSim353 \citep{finkelstein2001placing}, and SimVerb3500 \citep{gerz-etal-2016-simverb}. The evaluation metric is the Spearman correlation between the predicted cosine similarity vector and the ground truth.\\

\noindent \textbf {Extrinsic Evaluators:} We evaluate on the semantic textual similarity benchmarks (STS) from year $2012$ to $2016$ using SentEval \citep{conneau-kiela-2018-senteval}. Here, the task is to measure the semantic equivalence of a pair of sentences solely based on their cosine coefficient. We are particularly interested in these benchmarks for two reasons. 1) They evaluate the generalization power of the given vector space without any fine-tuning. 
2) Since they contain sentences from various sources such as news headlines and public forums, they reveal whether abstract knowledge is still preserved by our framework. We used BoW (averaging) to obtain sentence representations. While BoW is a simple sentence encoder, it is a great tool to evaluate the underlying structure of a vector space. For instance, the BoW representation of a pair of sentences such as `her dog is very smart' and `his cat is too dumb' are, unfortunately, very similar in a vector space that does not distinguish dissimilar from related words (e.g., smart and dumb). We will show that our model properly refines the textual vector space and alleviates these kinds of irregularities.


\begin{table}[ht]
\centering
\begin{adjustbox}{width=\columnwidth}
\begin{tabular}{lccccccccc}
\hline \textbf{Model} & \textbf{RW} & \textbf{MEN} & \textbf{WSim} & \textbf{MTurk} & \textbf{SimVerb} & \textbf{SimLex} & \textbf{Mean}\\
&  &  & \textbf{353} & \textbf{771} & \textbf{3500} & \textbf{999} & \\ \hline

GloVe   & 45.5 & 80.5 & 73.8 & 71.5 & 28.3 & 40.8 & 56.7\\
C\_GloVe   & 46 & 82.1 & 74.1 & 72.3 & 29.3 & 43.3 & 57.85\\
VGE\_G &  \bftab 52.6 & \bftab 85.1 & \bftab 78.9 & \bftab 73.4 &  \bftab 37.4 & \bftab 51.8 & \bftab 63.2\\
\hline
FastText     & 56.1 & 81.5 & 72.2 & 75.1 & 37.8 & 47.1 & 61.6\\
C\_fastText     & 49.2 & 68.3 & 58.1 & 56.8 & 30.3 & 41.9 & 50.76\\
VGE\_F & \bftab 57.8 & \bftab 83.6 &  \bftab 73.9 & \bftab 76.1 & \bftab 39.2 & \bftab 49.0 & \bftab 63.2\\


\hline

\end{tabular}
\end{adjustbox}
\caption{\label{intrinsic}  
Intrinsic evaluation. Visual grounding (denoted by `VGE') improves the results compared to the baselines on all test sets. 
}
\end{table}


\begin{table}[ht]
\centering
\begin{adjustbox}{width=\columnwidth}
\begin{tabular}{lcccccccc}
\hline \textbf{Model} & \textbf{RW} & \textbf{MEN} & \textbf{WSim} & \textbf{MTurk} & \textbf{SimVerb} & \textbf{SimLex}\\
&  &  & \textbf{353} & \textbf{771} & \textbf{3500} & \textbf{999} & \\ \hline

VGE\_G &  52.6 & \bftab 85.1 & \bftab 78.9 &  73.4 &  37.4 & 51.8 \\
VGE\_F & \bftab 57.8 &  83.6 &  73.9 & \bftab 76.1 & \bftab 39.2 &  49.0 \\
Cap2Both     & 48.7 & 81.9 & 71.2 & \_ &\_ & 46.7 \\
Cap2Img      & 52.3 & 84.5 & 75.3 & \_ &\_ & 51.5 \\
Park et al.  & \_ & 83.8 &77.5 &\_ &\_ & \bftab 58.0 \\
Park\_VG.  & \_ & \_ & \_ &\_ &\_ & 15.7 \\
Collell et al.           & \_ & 81.3 & \_ &\_ &28.6 & 41.0 \\

\end{tabular}
\end{adjustbox}
\caption{\label{intrinsic_comp}Comparison of grounded embeddings to previous work on intrinsic tasks. Ours are denoted by VGE.}

\end{table}

\section{Results}
\label{sec:results}
\noindent \textbf{Intrinsic Evaluation -- Baselines:} Table~\ref{intrinsic} shows the intrinsic evaluation results for the baselines and our visually grounded embeddings (VGE\_F and VGE\_G for visually grounded fastText and Glove respectively). In general, fastText performs better on word-level tasks compared to GloVe, probably because it provides more context for each word by leveraging from its sub-words. The results also validate the efficacy of our proposed model since updating the embeddings on captions alone (C\_fastText and C\_GloVe) brings subtle or no improvements. By the proposed visual grounding, significant improvements are achieved on \emph{all} datasets for both fastText and GloVe.
Analyzing why the improvement varies across different datasets is difficult. However, the table reveals interesting properties. For instance, the improvement on SimLex999,  which focuses more on the similarity between words, is larger than that on WSim353, which does not distinguish between similarity and relatedness. Hence, visual grounding seems to prioritize similarity over relatedness. Considering the overall performance, it enhances both embeddings to the same level despite their fundamental differences.

\noindent \textbf{Intrinsic Evaluation -- Grounded Embeddings:} We compare our model to related grounded embeddings by \cite{collell2017imagined, park-myaeng-2017-computational, kiros-etal-2018-illustrative, kiela-etal-2018-learning} (Table~\ref{intrinsic_comp}).  We limit our comparison to those who adopted the pre-trained GloVe or fastText since these pre-trained models alone outperform many visually grounded embeddings such as \citep{hasegawa-etal-2017-incorporating, zablocki2017learning} on many of our evaluation datasets.

Conceptually, \citet{kiela-etal-2018-learning} also induces visual grounding on GloVe by using the MSCOCO data set. Even though they propose a number of tasks for training (Cap2Img: predicting the image vector from its caption, Cap2Cap: generate an alternative caption of the same image; Cap2Both: training by Cap2Cap and Cap2Img simultaneously) our model clearly outperforms them as ours integrate visual information without degraded performance on abstract words. 

\citet{park-myaeng-2017-computational} proposed a polymodal approach by creating and combining six different types of embeddings (linear and syntactic contexts, cognition, sentiment, emotion, and perception) for each word.
Even though they used two pre-trained embeddings (GloVe and Word2vec) and other resources,
our model still outperforms their approach on MEN and WSim353, but their approach is better on Simlex999. This performance can be attributed to the many-modality training as using only their visually grounded embeddings (Park\_VG) performs much worse. This clearly shows that their visual embeddings do not benefit abstract words  \citep[cf.][]{park-myaeng-2017-computational}. 
In summary, our approach benefits from capturing different perspectives of the words' meanings by learning the reversible mapping in the context of multi-task learning.


\begin{table*}[ht]
\centering
\begin{adjustbox}{width=.8\textwidth}
\begin{tabular}{lc|cccccccc}
\hline \textbf{Model} & \textbf{All} & \textbf{Adjs} & \textbf{Nouns} & \textbf{Verbs} & \textbf{Conc-q1} & \textbf{Conc-q2} & \textbf{Conc-q3} & \textbf{Conc-q4} & \textbf{Hard}\\ \hline

GloVe   & 40.8 & 62.2 & 42.8 & 19.6 & 43.3 & 41.6 & 42.3 & 40.2 & 27.2 \\
VGE\_G (ours) & \bftab 51.8 & \bftab 72.1 & 52.0   &\bftab 35   & \bftab 53.1 &\bftab 54.8 & 47.4 & 56.8 & \bftab 38.3\\
Picturebook &37.3& 11.7 & 48.2   & 17.3   & 14.4 &27.5 &  46.2 & \bftab 60.7 & 28.8\\
Picturebook+GloVe &45.5& 46.2 & \bftab 52.1   & 22.8   & 36.7 &41.7 &  \bftab 50.4 &  57.3 & 32.5\\
\hline
\end{tabular}
\end{adjustbox}
\caption{\label{tab:simlex} SimLex999 (Spearman's $\rho$) results. 
Conc-q1 and Conc-q4 contain the most abstract and concrete words respectively. Our embeddings (VGE\_G) generalize across different word types and strongly outperform all the others on most of the categories.}
\end{table*}

\noindent \textbf{Fine-Grained Intrinsic Evaluation:} we further evaluate our model on the different categories of SimLex999 divided into nine sections: all (the whole dataset), adjectives, nouns, verbs, concreteness quartiles (from $1$ to $4$ increasing the degree of concreteness), and hard pairs. The hard section indicates $333$ pairs whose similarity is hard to discriminate from relatedness. 
The results for our best embeddings on SimLex999 (VGE\_G) are shown in Table~\ref{tab:simlex}. We see a large improvement over GloVe in all categories. Some previous approaches such as \citep{park-myaeng-2017-computational} concluded that perceptual information would be beneficial only to concrete words (e.g., apple,  table) and would adversely affect abstract words (e.g., happy, freedom). However, our model succeeds in maintaining high-precision  co-occurrence statistics from the textual model while augmenting these with perceptual information, in such a way that the representations for abstract words are actually enhanced. Therefore, it outperforms GloVe not only on concrete pairs (conc-q4) but also on highly abstract pairs (conc-q1).\\
We compared the results on SimLex999 with another recent visually grounded model called Picturebook \citep{kiros-etal-2018-illustrative}, which employs a multi-modal gating mechanism (similar to a LSTM and GRU update gate) to fuse the Glove and Picturebook embeddings (Table~\ref{tab:simlex}). It uses image feature vectors pre-trained on a fine-grained similarity task with 100+ million images \citep{wang2014learning}. 
Picturebook's performance is highly biased toward concrete words (conc-q3, conc-q4) and performs worse than GloVe by nearly $29\%$ on highly abstract words (conc-q1). Picturebook + GloVe on the other hand shows better results but still performs worse on highly abstract words and adjectives. Our model (VGE\_G) can generalize across different categories and outperforms Picturebook+Glove with a large margin on most of the categories while being quite comparable on the others.

\begin{table}[h]
\begin{adjustbox}{width=\columnwidth}
\begin{tabular}{lcccccc}
\hline
\textbf{Model}       & \textbf{STS12}          & \textbf{STS13}          & \textbf{STS14}          & \textbf{STS15}          & \textbf{STS16}   & \textbf{Mean}         \\ \hline
GloVe       &52.25&	49.59&	54.72&	56.25&	51.39&	52.84\\
C\_GloVe       & 53.27          & 50.56         &   56.72        &   57.86        & 52.11   &54.10  \\
VGE\_G    & \textbf{55.31} &	\textbf{57.24} &	\textbf{65.54}&	\textbf{67.61} &	\textbf{65.87} &	\textbf{62.35} \\ \hline
Fasttext    & 22.95&	24.63&	31.37&	37.71&	29.34&	29.2\\ 
C\_fasttext    &   29.69      &  23.80       &  37.58         &   45.29       & 29.34    & 33.14\\ 
VGE\_F & \textbf{31.78} & \textbf{32.26} &	\textbf{42.51}&	\textbf{48.79}&	\textbf{38.15}&	\textbf{38.70}\\ 

\hline
VGE\_G (ours)  & \textbf{55} & 57 & \textbf{66} &	\textbf{68} &66 & \textbf{62.40} \\
Word2vec    &52 &	\textbf{58} &	\textbf{66}& 	\textbf{68}&	65&	61.80\\
ELMo (top\_layer) & 54&	49&	62&	67&	63&	59.00\\
ELMo (all\_layers) & \textbf{55}&	51&	63&	69&	64&	60.40\\
Power-mean & 54 &	52&	63 &	66&	\textbf{67}&	60.40\\
\hline
\end{tabular}
\end{adjustbox}
\caption{\label{tab:STS} Comparison (Pearson correlation $\times 100$) of our embeddings (VGE\_*) with baselines (first two sections) and other word embeddings (bottom) on STS.}
\end{table}

\begin{table*}[ht]
\centering
\begin{adjustbox}{width=1\textwidth}


\begin{tabular}{cccccccccccccc}
\hline
\multicolumn{2}{c|}{happy} & \multicolumn{2}{|c|}{sad}  & \multicolumn{2}{|c|}{big} & \multicolumn{2}{|c|}{bird} & \multicolumn{2}{|c|}{horse} & \multicolumn{2}{|c|}{together} & \multicolumn{2}{|c}{smart}    \\ \hline
G           & V           & G        & V             & G         & V           & G           & V          & G        & V              & G           & V              & G             & V            \\
lucky       & pleased     & sadly    & saddened      & hard      & humongous   & turtle      & sparrow    & dog      & racehorse      & well        & togeather      & sensible      & witty        \\
everyone    & delighted   & shame    & tragic        & little    & Big         & nest        & Birds      & riding   & Thoroughbred   & bring       & togheter       & dumb          & shrewd       \\
love        & merry       & horrible & mournful      &           &             & squirrel    & avian      & ponies   & Horses         & both        & toegther       & sophisticated & inteligent   \\
always      & thrilled    & scared   & saddening     &           &             &             &            & donkey   & steed          & they        & togather       & attractive    & resourceful  \\
wish        & joyful      & awful    & sorrowful     &           &             &             &            &          &                & apart       & togethor       & wise          & quick-witted \\
hope        & hapy       & pity     & Sad           &           &             &             &            &          &                & up          & 2gether        &               &              \\
            &             & kinda    & heartbreaking &           &             &             &            &          &                & them        & togehter       &               &              \\
            &             & sorry    & heartbroken   &           &             &             &            &          &                & put         & togther        &               &              \\
            &             &          &               &           &             &             &            &          &                & along       & toghether      &               &              \\
            &             &          &               &           &             &             &            &          &                & with        & gether         &               &              \\
            &             &          &               &           &             &             &            &          &                &             &                &               &              \\ \hline
\end{tabular}

\end{adjustbox}
\caption{\label{tab:NN} Results of 10 nearest neighbors for Glove (G) and VGE\_G (V). Only the differing neighbors are reported. While GloVe retrieves more related words, ours (VGE\_G) focuses on similar words. Overall, VGE\_G is closer to human judgment and retrieves highly semantically similar words.}
\end{table*}

\noindent \textbf{Refining the Textual Vector Space:} Our grounded embeddings, while improving relatedness scores, prioritize similarity over relatedness. This is further demonstrated through inspection of nearest neighbors (Table~\ref{tab:NN}). 
Given the word `bird', GloVe returns `turtle' and `nest' while grounded GloVe returns `sparrow' and `avian', which both reference birds. 
Moreover, our embeddings retrieve more meaningful words regardless of the degree of abstractness. For the word `happy' for example, GloVe suffers from a bias toward dissimilar words with high co-occurrence such as `everyone', `always', and `wish'.  This issue is intrinsic to the fundamental assumption of the distributional hypothesis that words in the same context tend to be semantically related. Therefore, Glove embeddings, even though trained on 840 billion tokens, still reports antonyms such as `smart' and `dumb' as very similar. In addition, common misspellings of words (e.g., `togther') while serving the same role, occur with different frequencies in changing context. 
Hence, they are pulled apart in purely text-based vector spaces. However, our visual grounding model clearly puts them in the same cluster. Our model therefore seems to refine the text-based vector space by aligning it (via the mapping matrix) with real-world relations (in the images). This refinement generalizes to all the words by using our zero-shot mapping matrix which explains the improvement on highly abstract words. A sample of nearest neighbors for FastText and VGE\_F is available in Appendix~\ref{sec:appendix_fasttext_NN}. However, since FastText already performs quite well on intrinsic tasks, the difference with its grounded version is subtle which also confirms the results in Table~\ref{intrinsic}.

\noindent \textbf{Extrinsic Evaluation:} Table~\ref{tab:STS} shows the results on semantic similarity benchmarks. Both grounded embeddings strongly outperform their textual version on \emph{all} benchmarks. While fastText outperforms GloVe on intrinsic tasks, GloVe is superior here. The reason might be that unlike fastText GloVe treats each word as a single unit and takes into account the global co-occurrences of words.  This probably helps to capture the high-level structure of words (e.g., in sentences). Considering the mean score, our model boosts both embeddings approximately by $10$ percent.\\ 
Furthermore, while we are well aware that our simple averaging model cannot compete with the state-of-the-art sequence models~\citep{gao2021simcse} on the sentence level STS task, we compare it to other word embeddings to highlight the contribution of visual grounding. 
Table~\ref{tab:STS} (bottom) shows the results of our best model (VGE\_G) with other textual word embeddings namely ELMo \citep{peters2018deep}, Word2Vec \citep{mikolov2013distributed}, and Power-Mean \citep{ruckle2018concatenated} reported by \citep{perone2018evaluation}. While the textual GloVe is the second-worst model (by mean score: 52.84) in the table, its grounded version VGE\_G is the best one. 
Overall, these results confirm that 1) our grounding framework effectively integrates perceptual knowledge that is missing in purely text-based embeddings and 2) visual grounding is highly beneficial for downstream language tasks. 
It would be interesting to see if our findings extend to grounded sentence embedding models \citep{sileo2021visual,bordes-etal-2019-incorporating,tan2020vokenization} for instance by training transformer-based models such as BERT \citep{devlin2018bert} on top of our embeddings. However, we postpone this to the future since our focus here is on grounding word embeddings.

\begin{table}
\centering
\begin{adjustbox}{width=\columnwidth}

\begin{tabular}{llcc}
\hline \textbf{Dataset} & \textbf{Best $\alpha$} & \textbf{ Acc. with best $\alpha$}& \textbf{Acc. with $\alpha=1$} \\ \hline

RareWords       & 1.00&52.6 & 52.6 \\
MEN             & 0.63& 85.2 & 85.1 \\
WSim353           & 0.57& 79.3 & 78.9 \\
Mturk771       & 0.52& 74.2 & 73.4 \\
SimVerb3500     & 1.00& 37.4 & 37.4 \\
SimLex999       & 1.00& 51.8 & 51.8 \\

\hline
\end{tabular}
\end{adjustbox}
\caption{\label{tab:sensitivity} Sensitivity analysis (Spearman's $\rho$) on intrinsic datasets. $\alpha =1$ indicates no use of GloVe and $\alpha =0$ means no use of VGE\_G. Pure grounded embeddings alone yield the best results on 3 of the datasets.}
\end{table}

\begin{table}
\centering
\begin{adjustbox}{width=\columnwidth}

\begin{tabular}{lccccc}
\hline \textbf{Embeddings} &$\mathbf{\mathcal{L}_{FW}}$ & $\mathbf{\mathcal{L}_{FW} + \mathcal{L}_{BW} }$ & $\mathbf{\mathcal{L}_{FW} + \mathcal{L}_{BW} + \mathcal{L}_{B} }$& $\mathbf{\mathcal{L}_{All} + \mathcal{R}(\alpha,\beta) }$ \\ \hline

VGE\_G       & 61.60 & 61.82 & 62.66 & 63.20  \\
VGE\_F             & 61.70 & 61.83 & 61.60 & 63.20\\

\hline
\end{tabular}
\end{adjustbox}
\caption{\label{tab:ablation} Mean score (Spearman's $\rho$) on intrinsic datasets with respect to each task.  $\mathcal{L}_{All}$ refers to all the three tasks and $\mathcal{R}(\alpha,\beta)$ the regularization loss. }
\end{table}

\section{Model Analysis}
We further analyze the performance of our model from different perspectives as follows.\\ \textbf{Dependency on the Encoding Dimension $\mathbf{c}$:} 
We train our model with different dimensions of the grounded embeddings and measure the mean accuracy of all the intrinsic datasets. Table~\ref{tab:size_effect} shows the results using GloVe and VGE\_G with different sizes. Significant improvement is already achieved keeping the original dimension of GloVe (300). Higher dimensions up to a certain threshold ($1024$) increase the accuracy but beyond this point, the model starts to overfit.

\noindent \textbf{Dependency on the Textual Embeddings:} Further, we analyze how much of GloVe's original properties are maintained by the visual grounding.
Given $V_w$ and $G_w$ as the VGE\_G and GloVe vectors for the word $w$, we create a vector containing both embeddings \(E_w = [(1-\alpha)G_w; \alpha V_w] \).
Varying the relative weight $\alpha \in (0,1]$ we evaluate on the intrinsic datasets in Table~\ref{tab:sensitivity}. 
Three of the datasets yield the best results using only the grounded embeddings. The reduction in accuracy regarding `MEN' is also very subtle. 
On `WSim353' and `Mturk771', however, the best results are achieved with $\alpha \approx 0.5$. This might be because these datasets focus on the relatedness of words
while SimLex999 for instance distinguishes between similarity and relatedness.

\noindent \textbf{Ablation Study:} We further analyze the contribution of each task by performing an ablation evaluation. Table~\ref{tab:ablation} shows the mean score on all the intrinsic datasets (see Table~\ref{intrinsic}) with respect to each loss for both embeddings. While both GloVe and FastText show the same behaviour for language model tasks, fastText embeddings require more deviation ($\beta = 0$ in $\mathcal{R}(\alpha,\beta)$) to adapt to the binary discrimination task ($\mathcal{L}_{BW}$). Textual embeddings $T_e$ were frozen for all the cases except for $\mathcal{L}_{All}$. Even though the best performance, considering all the datasets, is achieved by using all the losses (including the regularization), each loss contributes differently to the overall performance. A more detailed ablation study based on the SimLex999 dataset is provided in Appendix~\ref{sec:appendix_ablation}.\\
\noindent \textbf{Connections to Human Cognition:}
Motivated by the different processing patterns of abstract and concrete words in the brain \citet{montefinese2019semantic}, we showed that it is possible to benefit from visual information without learning the two modalities in a joined space. Our experiments show that leveraging visual knowledge to inform the distributional models about the real world might be a better way of integrating language and vision. These modalities while separated could be informed and aligned with each other.

\begin{table}
\centering
\begin{adjustbox}{width=\columnwidth}

\begin{tabular}{l|cccccc}
\hline \textbf{Model\_dimensions} & \textbf{G\_300} & \textbf{V\_300} &\textbf{V\_512} &\textbf{V\_800} & \textbf{V\_1024} & \textbf{V\_2048}\\ \hline
\textbf{Mean Score} &  56.7 & 62.4 & 62.6 & 63.1 & \bftab 63.2 & 62.5 \\
\hline
\end{tabular}
\end{adjustbox}
\caption{\label{tab:size_effect} Effect of grounded word-vectors magnitude on intrinsic tasks. `G' and `V' refers to Glove and VGE\_G respectively. Significant improvement is achieved even with the same size as the textual GloVe.}

\end{table}

\section{Conclusion}

We investigated the effect of integrating perceptual knowledge from images into word embeddings via multi-task training. We constructed the visually grounded versions of GloVe and fastText by learning a zero-shot transformation from textual to grounded space trained on the MSCOCO dataset. Results on intrinsic and extrinsic evaluation support that visual grounding benefits current textual word embedding models. The major findings in our experiments are as follows:\\
a) Our improvement of visual grounding is not limited to words with concrete meanings and covers highly abstract words as well.\\
b) Discrimination between relatedness and similarity is more precise when using grounded embeddings.\\
c) Perceptual knowledge can profitably be transferred to purely textual downstream tasks.\\

Moreover, we showed that visual grounding has the potential to refine the irregularities in textual vector spaces by aligning words with their real-world relations.  This paves the way for future research on how visual grounding could resolve the problem of dissimilar words that occur frequently in the same context (e.g., small and big). In the future, we will investigate whether transformer blocks could profitably replace the GRU cells since they lead the state-of-the-art in many downstream sentence tasks. Moreover, while thus far our focus has been on words, a similar approach could be extended to obtain grounded sentence representations.

\section*{Acknowledgements}
This work has been supported by EXC number 2064/1 – Project number 390727645, as well as by the German Federal Ministry of Education and Research (BMBF): Tübingen AI Center, FKZ: 01IS18039A. The authors thank the International Max Planck Research School for Intelligent Systems (IMPRS-IS) for supporting Hassan Shahmohammadi. The third author was supported by ERC-WIDE (European Research Council -- Wide Incremental learning with Discrimination nEtworks), grant number 742545.

\clearpage
\bibliography{anthology,custom}
\bibliographystyle{acl_natbib}

\begin{table*}[t]
\centering


\resizebox{\textwidth}{!}{
\begin{tabular}{cccccccccccccc}
\hline

\multicolumn{2}{c|}{democracy} & \multicolumn{2}{|c|}{possible}  & \multicolumn{2}{|c|}{excited} & \multicolumn{2}{|c|}{round} & \multicolumn{2}{|c|}{medicine} & \multicolumn{2}{|c|}{flawlessly} & \multicolumn{2}{|c}{arrogantly}    \\ \hline
F           & V              & F             & V             & F         & V           & F           & V          & F              & V                 & F           & V              & F             & V            \\
dictatorship & democracy.    & necessary     &possibile   &  excitied  & EXCITED   & round.And      & round.It    & medical        & medecine          & flawless     & Flawlessly      & foolishly      & haughtily        \\
             &               & impossible    & possible.So & anxious   & excited-  & rounded       & round.The    & pharmacology   & pharmaceuticals   &        &        &           &        \\
            &                &               &             &           &           & oval           & -round      & medication     & medcine           &         &         &   &     \\
            &                &               &             &           &           & roundin        &  round.Now  &                &                  &          &         &      &    \\
 \hline
 
\end{tabular} }
 \caption{\label{tab:fasttext_NN} Results of 10 nearest neighbors for FastText(F) and VGE\_F (V). Only the differing neighbors are reported.}
  
\end{table*}

\newpage
\section{Appendix}
\appendix

\begin{table}[t]
\centering
\resizebox{\columnwidth}{!}{\begin{tabular}{l|c|cccccccc}
\hline \textbf{Model}  & \textbf{SimLex999} & \textbf{Adjs} & \textbf{Nouns} & \textbf{Verbs} & \textbf{Conc-q1} & \textbf{Conc-q2} & \textbf{Conc-q3} & \textbf{Conc-q4} & \textbf{Hard}\\ \hline

GloVe   & 40.8 & 62.2 & 42.8 & 19.6 & 43.3 & 41.6 & 42.3 & 40.2 & 27.2 \\
$\mathbf{\mathcal{L}_{B}}$  & 42.5 & 70.1 & 41.3& 25.1 & 45.8 & 45.9 & 43.8 & 46.6 &28.1 \\
$\mathbf{\mathcal{L}_{FW}}$ & \bftab 52.6 & 70.1 &53.1 & 37.8 & 54.4 & 54 & \bftab 49.3 & 55.2 &38 \\
$\mathbf{\mathcal{L}_{FW} + \mathcal{L}_{BW} }$ & 52.5 & 69.7 &52.6 & \bftab 40.6 & \bftab 55.5 & 54.1 & 48.7 & 55.4 &38.3 \\
$\mathbf{\mathcal{L}_{FW} + \mathcal{L}_{BW} + \mathcal{L}_{B} }$ & 52.5 & 69.8 & \bftab 53.5 & 37.7 &53.3 & 53.8 & 48.7 & \bftab 58 & \bftab 39.3 \\

$\mathcal{L}_{All} +\mathcal{R}(\alpha,\beta)  $ & 51.8 & \bftab 72.1 & 52.0   & 35   &  53.1 &\bftab 54.8 & 47.4 & 56.8 &  38.3\\

\hline
Fasttext   & 47.1 & 59.8 & 50.5 & 31.5 & 46.4 & \bftab 46.8 & 48.5 & 52 & 29.6 \\
$\mathbf{\mathcal{L}_{B}}$ & 38.5  &\bftab  64.9  &41.7 &23.8  &37.2  &37  &41.2  & 48 &26.2 \\
$\mathbf{\mathcal{L}_{FW}}$ & 50.2 & 59 & 55.8 & \bftab 37.1 & 47.1 & 46.1 & 51.9 & 60.2 & 32.1 \\
$\mathbf{\mathcal{L}_{FW} + \mathcal{L}_{BW} }$  & \bftab 50.8  & 59.3 & \bftab 57.3 & 36 & 46.1 & 46.2 & \bftab 53.1 & \bftab 62.3 &32.7 \\
$\mathbf{\mathcal{L}_{FW} + \mathcal{L}_{BW} + \mathcal{L}_{B} }$  &  \bftab  50.8 & 60.6 & 57.1   & 35.8   & \bftab 48.1 & 46.4 & 52.7 & 61.3 & \bftab 33.4\\
$\mathcal{L}_{All} +\mathcal{R}(\alpha,\beta)  $  &  49.0 &  58.6 & 54.1   & 32.9   &  45.3 & 46.7 & 51.3 & 57.7 &  31.3\\

\hline
\end{tabular}}
\caption{\label{tab:ablation_fine_grained} Fine-grained ablation study on SimLex999 (Spearman's $\rho$). Conc-q1 and Conc-q4 contain the most abstract and concrete words respectively. The hard section includes a set of word-pairs in which similarity is hard to distinguish from relatedness}
\end{table}

\section{Fine-Grained Ablation Study}
\label{sec:appendix_ablation}
In this section, we provide a more detailed ablation study based on the SimLex999 dataset for both FastText and GloVe. Shown in Table~\ref{tab:ablation_fine_grained}, the results reveal interesting findings. The binary discrimination task ($\mathbf{\mathcal{L}_{B}}$) is the most beneficial one for adjectives in the case of both embeddings. This improvement arguably comes from the missing information in textual representations such as shapes, colors, and sizes of the objects which are fused by this cross-modality alignment. $\mathbf{\mathcal{L}_{B}}$ also boosts the performance of the `Hard' section in which similarity is hard to distinguish from relatedness. The reason probably lies in the shift of focus toward similarity (see Table~\ref{tab:NN}) which makes it easier to distinguish between similarity and relatedness. The language model tasks ($\mathbf{\mathcal{L}_{FW}}$ and $\mathbf{\mathcal{L}_{BW}}$) seem to contribute the most to nouns and verbs describing the scenes in the images. Moreover, our best model ($\mathbf{\mathcal{L}_{All} + \mathcal{R}(\alpha,\beta) }$), regarding all the datasets, does not achieve the best result here because each dataset focuses on a different aspect of the language (e.g, similarity or relatedness). However, our final embeddings incorporate the information from different perspectives and improve on all the datasets.

\section{Refining   the   Textual   Vector   Space}
\label{sec:appendix_fasttext_NN}
Similar to the visually grounded GloVe embeddings, the grounded FastText (VGE\_F) also refine the irregularities of textual vector space (referring to Section~\ref{sec:results}). Examples of differing nearest neighbors are reported in Table~\ref{tab:fasttext_NN}. Since FastText performs quite well on word-level tasks, the difference is very subtle. The improvement seems to mainly fall into alleviating the antonym problem (e.g, for `democracy' in the table) and clustering typos together (e.g, `medicine' and `medecine'). We can also observe tokens such as `round.And' that FastText's tokenizer has failed to split but have been cluster together by our approach.  Overall, the table confirms the results in Table~\ref{intrinsic}.

\end{document}